\documentclass[runningheads]{llncs}
\usepackage{graphicx}
\usepackage{float} 
\usepackage{comment}
\usepackage{url}
\usepackage{todonotes}
\usepackage{hyperref}
\usepackage{multirow}

\begin{document}
\title{Automated Mining of \textsc{Leaderboards} for Empirical AI Research} 
%
%

%
%
\author{Salomon Kabongo\inst{1}\orcidID{0000-0002-0021-9729} \and
Jennifer D’Souza\inst{2}\orcidID{0000-0002-6616-9509} \and
Sören Auer\inst{1,2}\orcidID{0000-0002-0698-2864}}
\authorrunning{S. Kabongo et al.}

\institute{L3S Research Center, Leibniz University of Hannover, Hannover, Germany \email{kabenamualu@l3s.de} \and
TIB Leibniz Information Centre for Science and Technology
Hannover, Germany \email{\{jennifer.dsouza, soeren.auer\}@tib.eu}}

%
\maketitle              
\begin{abstract}
With the rapid growth of research publications, empowering scientists to keep oversight over the scientific progress is of paramount importance. In this regard, the \textsc{Leaderboards} facet of information organization provides an overview on the state-of-the-art by aggregating empirical results from various studies addressing the same research challenge. Crowdsourcing efforts like PapersWithCode among others are devoted to the construction of \textsc{Leaderboards} predominantly for various subdomains in Artificial Intelligence. Leaderboards provide machine-readable scholarly knowledge that has proven to be directly useful for scientists to keep track of research progress. The construction of \textsc{Leaderboards} could be greatly expedited with automated text mining.

This study presents a comprehensive approach for generating \textsc{Leaderboards} for knowledge-graph-based scholarly information organization. Specifically, we investigate the problem of automated \textsc{Leaderboard} construction using state-of-the-art transformer models, viz. Bert, SciBert, and XLNet. Our analysis reveals an optimal approach that significantly outperforms existing baselines for the task with evaluation scores above 90\% in F1. This, in turn, offers new state-of-the-art results for \textsc{Leaderboard} extraction. As a result, a vast share of empirical AI research can be organized in the next-generation digital libraries as knowledge graphs.

\end{abstract}
\section{Introduction}
Our present rapidly amassing wealth of scholarly publications~\cite{jinha2010article} poses a crucial dilemma for the research community. A trend that is only further bolstered in a number of academic disciplines with the sharing of PDF preprints ahead (or even instead) of peer-reviewed publications~\cite{chiarelli2019accelerating}. The problem is: \textit{How to stay on-track with the past and the current rapid-evolving research progress?} In this era of the publications deluge~\cite{stm}, such a task is becoming increasingly infeasible even within one’s own narrow discipline. Thus, the need for novel technological infrastructures to support intelligent scholarly knowledge access models is only made more imminent. A viable solution to the dilemma is if \textit{research results} were made skimmable for the scientific community with the aid of advanced knowledge-based information access methods. This helps curtail the time-intensive and seemingly unnecessary cognitive labor that currently constitute the researcher’s task of searching just for the results information in full-text articles to track scholarly progress~\cite{auer_2018}. Thus, \textit{strategic reading} of scholarly knowledge focused on core aspects of research powered by machine learning may soon become essential for all users~\cite{renear2009strategic}. Since the current discourse-based form of the \textit{results} in a static PDF format do not support advanced computational processing, it would need to transition to truly digital formats (at least for some aspects of the research). 

In this regard, an area gaining traction is in empirical Artificial Intelligence (AI) research. There, \textsc{Leaderboard}s are being crowdsourced from scholarly articles as alternative machine-readable versions of performances of the AI systems. \textsc{Leaderboard}s typically show quantitative evaluation results (reported in scholarly articles) for defined machine learning tasks evaluated on standardized datasets using comparable metrics. As such, they are a key element for describing the state-of-the-art in certain fields and tracking its progress. Thus, empirical results can be benchmarked in online digital libraries. Some well-known initiatives that exist to this end are: \textsc{PapersWithCode.com}~\cite{PWC}, NLP-Progress~\cite{np}, AI-metrics~\cite{aim}, SQUaD explorer~\cite{squad-exp}, Reddit SOTA~\cite{reddit}, and the Open Research Knowledge Graph.\footnote{\url{https://www.orkg.org/orkg/benchmarks}}

Expecting scientists to alter their documentation habits to machine-readable versions rather than human-readable natural language is unrealistic, especially given that the benefits do not start to accrue until a critical mass of content is represented in this way. For this, the retrospective structuring from pre-existing PDF format of results is essential to build a credible knowledge base. Prospectively, machine learning can assist scientists to record their results in the \textsc{Leaderboard}s of next-generation digital libraries such as the Open Research Knowledge Graph (ORKG)~\cite{jaradeh2019open}. In our age of the ``deep learning tsunami,''~\cite{manning2015computational} there are many studies that have used neural network models to improve the construction of automated scholarly knowledge mining systems~\cite{luan2018multi,brack2020domain,anteghini2020scibert,jiang2020improving}. With the recent introduction of language modeling techniques such as transformers~\cite{vaswani2017attention}, the opportunity to obtain boosted machine learning systems is further accentuated. 

In this work, we empirically tackle the \textsc{Leaderboard} knowledge mining machine learning (ML) task via a detailed set of evaluations involving a large dataset and several ML models. The \textsc{Leaderboard} concept varies wrt. the domains or the captured data. Inspired by prior work~\cite{hou2019identification,jain2020scirex,mondal2021end}, we define a \textsc{Leaderboard} comprising the following three scientific concepts: 1. \textit{\underline{T}ask}, 2. \textit{\underline{D}ataset}, and 3. \textit{\underline{M}etric} (TDM). However, this base \textsc{Leaderboard} structure can be extended to include additional concepts such as method name, code links, etc. In this work, we restrict our evaluations to the core TDM triple. Thus, constructing a \textsc{Leaderboard} in our evaluations entails the extraction of all related TDM statements from an article. E.g., (Language Modeling, Penn Treebank, Test perplexity) is a \textsc{Leaderboard} triple of an article about the `Language Modeling' \textit{Task} on the `Penn Treebank' \textit{Dataset} in terms of the `Test perplexity' \textit{Metric}. Consequently, the construction of comparisons and visualizations over such machine-interpretable data can enable summarizing the performance of empirical findings across systems. 

While prior work~\cite{hou2019identification,mondal2021end} has already initiated the automated learning of \textsc{Leaderboard}s, these studies were mainly conducted under a single scenario, i.e. only one learning model was tested, and over a small dataset. For stakeholders in the Digital Library (DL) community interested in leveraging this model practically, natural questions may arise: \textit{Has the optimal learning scenario been tested? Would it work in the real-world setting of large amounts of data?} Thus, we note that it should be made possible to recommend a technique for knowledge organization services from observations based on our prior comprehensive empirical evaluations~\cite{jiang2020improving}. Our ultimate goal with this study is to help the DL stakeholders to select the optimal tool to implement knowledge-based scientific information flows w.r.t. \textsc{Leaderboard}s. To this end, we evaluated three state-of-art transformer models, viz. Bert, SciBert, and XLNet, each with their own respective unique strengths. The automatic extraction of \textsc{Leaderboard}s presents a challenging task because of the variability of its location within written research, lending credence to the creation of a human-in-the-loop model. Thus, our \textsc{Leaderboard} mining system will be prospectively effective as intelligent helpers in the crowdsourcing scenarios of structuring research contributions~\cite{oelen2021crowdsourcing} within knowledge-graph-based DL infrastructures. Our approach called ORKG-TDM is developed and integrated into the scholarly knowledge organization platform Open Research Knowledge Graph (ORKG)~\cite{jaradeh2019open}.

In summary, the contributions of our work are:
\begin{enumerate}
    \item we construct a large empirical corpus containing over 4,500 scholarly articles and \textsc{Leaderboard} TDM triples for training and testing \textsc{Leaderboard} extraction approaches; 
    \item we evaluate three different transformer model variants in experiments over the corpus and integrate these into the ORKG-TDM \textsc{Leaderboard} extraction platform;
    \item in a comprehensive empirical evaluation of ORKG-TDM we obtain scores of 93.0\% micro and 92.8\% macro F1 outperforming existing systems by over 20 points.
\end{enumerate}

To the best of our knowledge, our ORKG-TDM models obtain state-of-the-art results for \textsc{Leaderboard} extraction defined as (\textit{Task}, \textit{Dataset}, \textit{Metric}) triples extraction from empirical AI research articles. Thus ORKG-TDM can be readily leveraged within KG-based DLs and be used to comprehensively construct \textsc{Leaderboard}s with more concepts beyond the TDM triples. Our data\footnote{\url{http://doi.org/10.5281/zenodo.5105798}} and code is made publicly available.\footnote{\url{https://github.com/Kabongosalomon/task-dataset-metric-nli-extraction}}

\section{Related Work}
Organizing scholarly knowledge extracted from scientific articles in a Knowledge Graph, has been viewed from various Information Extraction (IE) perspectives.

\subsubsection{Digitalization based on Textual Content Mining.}

Building a scholarly knowledge graph with text mining involves two main tasks: 1. scientific term extraction and 2. extraction of scientific or semantic relations between the terms.

Addressing the first task, several dataset resources have been created with scientific term annotations and the term concept typing to foster the training of supervized machine learners. For instance, the ACL RD-TEC dataset~\cite{handschuh2014acl} annotates computational terminology in Computational Linguistics (CL) scholarly articles and categorizes them simply as \textit{technology} and \textit{non-technology} terms. The ScienceIE SemEval 2017 shared task~\cite{augenstein2017semeval} annotates the full text in articles from Computer Science, Material Sciences, and Physics domains for \textit{Process}, \textit{Task} and \textit{Material} types of keyphrases. SciERC~\cite{Luan2018MultiTaskIO} annotates articles from the machine learning domain with six concepts \textit{Task}, \textit{Method}, \textit{Metric}, \textit{Material}, \textit{Other-Scien\-tific\-Term} and \textit{Generic}. The STEM-ECR~\cite{lrec2020} corpus annotates \textit{Process}, \textit{Method}, \textit{Material}, and \textit{Data} concepts in article abstracts interdisciplinarily across ten STEM disciplines. 

For the identification of relations between scientific terms in the natural language processing (NLP) community, within the context of human annotations on the abstracts of scholarly articles~\cite{augenstein2017semeval,gabor2018semeval}, seven relation types between scientific terms have been studied. They are \textsc{Hyponym-Of}, \textsc{Part-Of}, \textsc{Usage}, \textsc{Compare}, \textsc{Conjunction}, \textsc{Feature-Of}, and \textsc{Result}. The annotations are in the form of generalized relation triples: $\langle$experiment$\rangle$ \textsc{Compare} $\langle$another experiment$\rangle$; $\langle$method$\rangle$ \textsc{Usage} $\langle$data$\rangle$; $\langle$method$\rangle$ \textsc{Usage} $\langle$research task$\rangle$. Since human language exhibits the paraphrasing phenomenon, identifying each specific relation between scientific concepts is impractical. In the framework of an automated pipeline for generating knowledge graphs over massive volumes of scholarly records, the task of classifying scientific relations (i.e., identify the appropriate relation type for each related concept pair from a set of predefined relations) is therefore indispensable. 

In other text mining initiatives, comprehensive knowledge mining themes are being defined on scholarly investigations. The recent NLPContributionGraph Shared Task~\cite{nlpcontributiongraph,ncg-final-dataset,ncg} released KG annotations of contributions including the facets of research problem, approach, experimental settings, and results, in an evaluation series that showed it a challenging task. Similarly, in the Life Sciences, comprehensive KGs from reports of biological assays, wet lab protocols and inorganic materials synthesis reactions and procedures~\cite{anteghini2020scibert,anteghini2020representing,chemrecipes,labprotocols,kuniyoshi2020annotating,mysore2019materials} are released as ontologized machine-interpretable formats for training machine readers.

\subsubsection{Digitalization based on Table Mining}

In the earlier subsection, we discussed information extraction models defined for retrieving the relevant structured information from the textual body of articles. Recent efforts are geared to mining information from the semi-structured format of information in articles as tables. Unlike the high performances seen in information mining systems applied to textual data, text mining performances over tables are relatively much lower.

Milosevic et al.~\cite{milosevic} tested methods for extracting numerical (number of patients, age, gender distribution) and textual (adverse reactions) information from tables identified by the $<table>$ tag in the clinical literature as XML articles. Further, another line of work examined the classification of tables from HTML pages as entity, relational, matrix, list, and nondata leveraging specialized table embeddings called TabVec~\cite{ghasemi2018tabvec}. Wei et al.\cite{weitabextract} defined a question answering task with data in Table cells as the answers over two different datasets, i.e. web data tables and news articles in text format tables. Another model called \textsc{TaPaS}~\cite{herzig2020tapas} also addressed question answering over tabular data by extending BERT’s architecture to encode tables as input and training it end-to-end over tables crawled from Wikipedia. TableSeer~\cite{liu2007tableseer} is a comprehensive tables mining search engine that crawls digital libraries, detects tables from documents, extracts their metadata, and indexes and ranks tables in a user-friendly search interface.

\subsubsection{Digitalization based on Textual and Tabulated Content Mining}

IBM's science result extractor~\cite{hou2019identification} first defined the $(Task,Dataset,Metric)$ extraction task from articles. They trained a BERT classification model leveraging context data from the abstract, tables, and from table headers and captions. Their dataset comprised pdf-to-text converted articles from \url{PapersWithCode.com}. Following which, AxCell~\cite{kardas2020axcell} presented an automated machine learning pipeline for extracting results from papers. It used several novel components, including table segmentation, to learn relevant structural knowledge to aid extraction. Unlike the first system, AxCell was trained and tested over \LaTeX ~source code of machine learning papers from \url{arXiv.org}. Furthermore, the SciRex corpus creation endeavor~\cite{jain2020scirex} defines mostly similar information targets as science result extractor. However, SciRex is evaluated on clean \LaTeX ~sources unlike the IBM extractor and our objective of trying to identify robust machine learning pipelines over articles in PDF format. Another recent system, SciNLP-KG~\cite{mondal2021end}, reformulated the \textsc{Leaderboard} extraction task as one with relation \textit{evaluatedOn} between tasks and datasets, \textit{evaluatedBy} between tasks and metrics, as well as coreferent and related relations between the same type of entities. Like us, they comprehensively investigated several transformer model variants. Nevertheless, owing to their task reformulation, relation-based evaluation, and different dataset our results cannot be compared. Finally, Hou et al., the developers of IBM's science result extractor, recently released the TDMSci corpus~\cite{hou2021tdmsci} as a sequence labeling task for extracting \textit{Task}, \textit{Dataset}, and \textit{Metric} at the sentence-level. We maintain the original document-level inference task definition.

In this paper, we investigate the science result extractor system~\cite{hou2019identification}, however, with a detailed empirical perspective. We comprehensively evaluate the potential of transformer models for the task by testing Bert, SciBert, and XLNet. We also test the models over a much larger empirical dataset emulating its application in practice in the framework of the scholarly digital libraries such as the ORKG.

\section{Our Leaderboards Labeled Corpus}

To facilitate supervised system development for the extraction of \textsc{Leaderboard}s from scholarly articles, we build an empirical corpus that encapsulates the task. \textsc{Leaderboard} extraction is essentially an inference task over the document. To alleviate the otherwise time-consuming and expensive corpus annotation task involving expert annotators, we leverage distant supervision from the available crowdsourced metadata in the PwC KB. In the remainder of this section, we explain our corpus creation and annotation process.

\subsubsection{Scholarly Papers and Metadata from the PwC Knowledge Base.}
We created a new corpus as a collection of scholarly papers with their TDM triple annotations for evaluating the \textsc{Leaderboard}s extraction task inspired by the original IBM science result extractor~\cite{hou2019identification} corpus. The collection of scholarly articles for defining our \textsc{Leaderboard} extraction objective is obtained from the publicly available crowdsourced leaderboards PapersWithCode (PwC)\footnote{\url{https://paperswithcode.com/}}. It predominantly represents articles in the Natural Language Processing and Computer Vision domains, among other AI domains such as Robotics, Graphs, Reasoning, etc. Thus, the corpus is representative for empirical AI research. The original downloaded collection (timestamp 2021-05-10 at 12:30:21)\footnote{Our corpus was downloaded from the PwC Github repository \url{https://github.com/paperswithcode/paperswithcode-data} and was constructed by combining the information in the files \textit{All papers with abstracts} and \textit{Evaluation tables} which included article urls and TDM crowdsourced annotation metadata.} was pre-processed to be ready for analysis. While we use the same method here as the science result extractor, our corpus is different in terms of both labels and size, i.e. number of papers, as many more \textsc{Leaderboard}s have been crowdsourced and added to PwC since the original work.

\subsubsection{PDF pre-processing}
While the respective articles' metadata in machine-readable form was directly obtained from the PwC data release, the document itself being in PDF format needed to undergo pre-processing for pdf-to-text conversion so that its contents could be mined. For this, the GROBID parser~\cite{lopez2009grobid} was applied to extract the title, abstract, and for each section, the section title and its corresponding content from the respective PDF article files. Each article's parsed text was then annotated with TDM triples via distant labeling to create the final corpus. 

\subsubsection{Paper Annotation via Distant Labeling}
Each paper was associated with its \textsc{Leaderboard} TDM triple annotations. These were available as the crowdsourced metadata of each article in the PwC knowledge base (KB). The number of triples per article varied between 1 (minimum) and 54 (maximum) at an average of 4.1 labels per paper. The corpus was thus annotated as a distant labeling task since the labels for each paper were directly imported from the PwC KB without additional human curation of the varying forms of label names. Additionally, \textsc{Leaderboard}s that appeared in less than five papers were ignored. Consequently to the TDM labels filtering stage, some articles were without TDM triples and these articles were annotated with the label ``unknown.'' 

Our overall corpus statistics are shown in \autoref{table:dataset_stats}. We adopted the 70/30 split for the Train/Test folds for the empirical system development (described in detail in \autoref{sec:mining}). In all, our corpus contained 5,361 articles split into 3,753 in the training data and 1,608 in the test data. There were  unique TDM-triples  overall. Note that since the test labels were a subset of the training labels, the unique labels overall can be considered as those in the training data. \autoref{table:dataset_stats} also shows the distinct \textit{Tasks}, \textit{Datasets}, \textit{Metrics} in the last three rows. Our corpus contains 288 \textit{Tasks} defined on 908 \textit{Datasets} and evaluated by 550 \textit{Metrics}. This is significantly larger than the original corpus which had 18 \textit{Tasks} defined on 44 \textit{Datasets} and evaluated by 31 \textit{Metrics}.

\begin{table}[!tb]
\centering  \scriptsize
\begin{tabular}{l|r|r|r|r|}
\multirow{2}{*}{} & \multicolumn{2}{c}{\textbf{Ours}} & \multicolumn{2}{|c|}{\textbf{Original}} \\
                  & Train       & Test       & Train       & Test  \\ \hline
Papers               & 3,753  &  1,608  & 170  &  167  \\
``unknown'' annotations &  922 & 380 &  46 &  45 \\
Total TDM-triples & 11,724 & 5,060 & 327  & 294 \\
Avg. number of TDM-triples per paper & 4.1 & 4.1 &  2.64 & 2.41 \\
Distinct TDM-triples & 1,806 & 1,548 & 78  & 78 \\
Distinct \textit{Tasks}       & 288 & 252 & 18  & 18 \\
Distinct \textit{Datasets}    & 908 & 798 &  44 & 44 \\
Distinct \textit{Metrics}     & 550 & 469 &  31 & 31 \\
\end{tabular}
\caption{Ours vs. the original science result extractor~\cite{hou2019identification} corpora statistics. The ``unknown'' labels were assigned to papers with no TDM-triples after the label filtering stage.}
\label{table:dataset_stats}
\end{table}

\begin{figure}[!tb]
	\centering
	\includegraphics[width=1.0\linewidth]{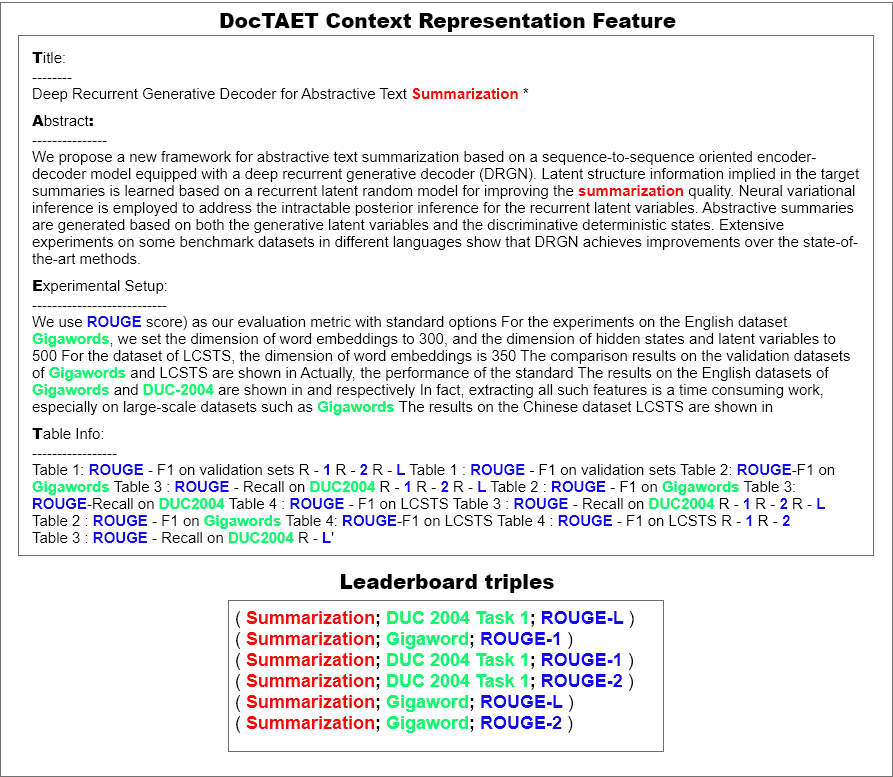}
	\caption{The DocTAET model with context features as a concatenation of the Scholarly \underline{\textbf{Doc}}ument's \underline{\textbf{T}}itle, \underline{\textbf{A}}bstract, \underline{\textbf{E}}xperimental-Setup, and \underline{\textbf{T}}able content/captions for training NLI transformer models on a set of \textsc{Leaderboard} triples. The figure illustrates specifically six (\textcolor{red}{\textit{task}}, \textcolor{green}{\textit{dataset}}, \textcolor{blue}{\textit{metric}}) triples and their context in the original article text extracted as the feature for transformer models.}
	\label{fig:f1}
\end{figure}

\section{Leaderboard Extraction Task Definition}
\label{sec:task}

The task is defined on the dataset described in previous section. The dataset can be formalized as follows. Let $p$ be a paper in the collection $P$. Each $p$ is annotated with at least one triple $(t_i,d_j,m_k)$ where $t_i$ is the $i^{th}$ task defined, $d_j$ the $j^{th}$ dataset and $m_k$ the $k^{th}$ system evaluation metric. The number of triples per paper vary.

In the supervised inference task, the input data instance corresponds to the pair: a paper $p$ represented as the DocTAET context feature $p_{DocTAET}$ and its TDM-triple $(t,d,m)$. The inference data instance, then is $(c; [(t,d,m), p_{DocTAET}])$ where $c \in \{true, false\}$ is the inference label. Thus, specifically, our \textsc{Leaderboard} extraction problem is formulated as a \textit{natural language inference} task between the DocTAET context feature $p_{DocTAET}$ and the $(t,d,m)$ triple annotation. $(t,d,m)$ is $true$ if it is among the paper's TDM-triples, otherwise $false$. The $false$ instances are artificially created by random selection of $(t,d,m)$ annotations from another paper. Cumulatively, \textsc{Leaderboard} construction is a multi-label, multi-class inference problem.

\subsection{DocTAET Context Feature}

In \autoref{fig:f1}, we depict the DocTAET context feature~\cite{hou2019identification}. Essentially, the \textsc{Leaderboard} extraction task is defined on the full document content. However, the respective (\textit{task}, \textit{dataset}, \textit{metric}) label annotations are mentioned only in specific places in the full paper such as in the Title, Abstract, Introduction, Tables. The DocTAET feature was thus defined to capture the targeted context information to facilitate the (\textit{task}, \textit{dataset}, \textit{metric}) triple inference. It focused on capturing the context from four specific places in the text-parsed article, i.e. from the title, abstract, first few lines of the experimental setup section as well as table content and captions.

\section{Transformer-based Leaderboard Extraction Models}

For \textsc{Leaderboard} extraction~\cite{hou2019identification}, we employ deep transfer learning modeling architectures that rely on a recently popularized neural architecture -– the transformer~\cite{vaswani2017attention}. Transformers are arguably the most important architecture for natural language processing (NLP) today since they have shown and continue to show impressive results in several NLP tasks~\cite{devlin2018bert}. Owing to the self-attention mechanism in the transformer models, they can be fine-tuned on many downstream tasks. These models have thus crucially popularized the transfer learning paradigm in NLP. We investigate three transformer-based model variants for \textsc{Leaderboard} extraction in a Natural Language Inference configuration. 

Natural language inference (NLI), generally, is the task of determining whether a ``hypothesis'' is true (entailment), false (contradiction), or undetermined (neutral) given a ``premise''~\cite{ref_paperswithcode_nli}. For \textsc{Leaderboard} extraction, the slightly adapted NLI task is to determine that the (\textit{task}, \textit{dataset}, \textit{metric}) ``hypothesis'' is true (entailed) or false (not entailed) for a paper given the ``premise'' as the DocTAET context feature representation of the paper.

Currently, there exist several transformer-based models. In our experiments, we investigated three core models: two variants of Bert, i.e. the vanilla Bert~\cite{devlin2018bert} and the scientific Bert (SciBert)~\cite{beltagy2019scibert}. We also tried a different type of transformer model than Bert called XLNet~\cite{yang2019xlnet} which employs Transformer-XL as the backbone model. Next, we briefly describe the three variants we use.

\subsection{Bert Models}
Bert (i.e., Bidirectional Encoder Representations from Transformers), is a bidirectional autoencoder (AE) language model. As a pre-trained language representation built on the deep neural technology of transformers, it provides NLP practitioners with high-quality language features from text data simply out-of-the-box and thus improves performance on many NLP tasks. These models return contextualized word embeddings that can be directly employed as features for downstream tasks~\cite{jiang2020improving}.

The first Bert model we employ is Bert\textsubscript{base} (12 layers, 12 attention heads, and 110 million parameters) which was pre-trained on billions of words from the BooksCorpus (800M words) and the English Wikipedia (2,500M words).

The second Bert model we employ is the pre-trained scientific Bert called SciBert~\cite{beltagy2019scibert}. SciBert was pretrained on a large corpus of scientific text. In particular, the pre-training corpus is a random sample of 1.14M papers from Semantic Scholar\footnote{\url{https://semanticscholar.org}} consisting of full texts of 18\% of the papers from the computer science domain and 82\% from the broad biomedical domain. For both Bert\textsubscript{base} and SciBert, we used their uncased variants.

\subsection{XLNet}

XLNet is an autoregressive (AR) language model~\cite{yang2019xlnet} that enables learning bidirectional contexts using Permutation Language Modeling, unlike Bert which uses Masked Language Modeling. Thus in PLM all tokens are predicted but in random order, whereas in MLM only the masked (15\%) tokens are predicted. This is also in contrast to the traditional language models, where all tokens were predicted in sequential order instead of random order. Random order prediction helps the model to learn bidirectional relationships and therefore better handle dependencies and relations between words. In addition, it uses Transformer XL~\cite{dai2019transformer} as the base architecture, which models long contexts unlike the Bert models with contexts limited to 512 tokens.
Since only cased models are available for XLNet, we used the cased XLNet\textsubscript{base} (12 layers, 12 attention heads, and 110 million parameters).

\section{Automated Leaderboard Mining}
\label{sec:mining}
\subsection{Experimental Setup}

\subsubsection{Parameter Tuning.}

We used the Hugging Transfomer libraries\footnote{\url{https://github.com/huggingface/transformers}} with their Bert variants and XLNet implementations. In addition to the standard fine-tuned setup for NLI, the transformer models were trained with a learning rate of $1e-5$ for 14 epochs; and used the $AdamW$ optimizer with a weight decay of 0 for \textit{bias}, \textit{gamma}, \textit{beta} and 0.01 for the others. Our models' hyperparameters details are available online.\footnote{\url{https://github.com/Kabongosalomon/task-dataset-metric-nli-extraction/blob/main/train_tdm.py}}

In addition, we introduce a task-specific parameter that was crucial in obtaining optimal task performance from the models. It was the number of \textit{false} triples per paper. This parameter controls the discriminatory ability of the model. The original science result extractor system~\cite{hou2019identification} considered $|n|-|t|$ \textit{false} instances for each paper, where $|n|$ was the distinct set of triples overall and $|t|$ was the number of \textit{true} \textsc{Leaderboard} triples per paper. This approach would not generalize to our larger corpus with over 2,500 distinct triples. In other words, considering that each paper had on average 4 \textit{true} triples, it would have 2,495 \textit{false} triples which would strongly bias the classifier learning toward only \textit{false} inferences. Thus, we tuned this parameter in a range of values in the set \{10, 50, 100\} which at each experiment run was fixed for all papers.

Finally, we imposed an artificial trimming of the DocTAET feature to accommodate the Bert models maximum token length of 512. For this, the token lengths of the experimental setup and table info were initially truncated to approximately $150$, after which the complete DocTAET feature is trimmed to 512 tokens.

\subsubsection{Two-Fold Cross Validation.} To evaluate robust models, we performed two-fold cross validation experiments. In each fold experiment, we train a model on 70\% of the overall dataset, and test on the remaining 30\% ensuring that the test data splits are not identical between the folds. Thus, all cumulative results reported are averaged results over the two folds. Also, \autoref{table:dataset_stats} corpus statistics are averaged estimates over the two experimental folds.

\subsubsection{Evaluation Metrics.} Within the two-fold experimental settings, we report macro- and micro-averaged precision, recall, and F1 scores for our \textsc{Leaderboard} extraction task on the test dataset. The macro scores capture the averaged class-level task evaluations, whereas the micro scores represent fine-grained instance-level task evaluations.

Further, the macro and micro evaluation metrics for the overall task have two evaluation settings: 1) considers papers with \textsc{Leaderboard}s and papers with ``unknown'' in the metric computations; 2) only papers with \textsc{Leaderboard}s are considered while the papers with ``unknown'' are excluded.

\begin{table}[!bt]
\centering  \scriptsize
\begin{tabular}{lcccccc}
\hline
\textbf{}        & \textbf{Macro P} & \textbf{Macro R} & \textbf{Macro F1} & \textbf{Micro P} & \textbf{Micro R} & \textbf{Micro F1} \\ 
\hline
\multicolumn{7}{c}{(a) Task + Dataset + Metric Extraction} \\ 
\hline
TDM-IE\textsubscript{Bert} & 62.5&  75.2 & 65.3 & 60.8 & {76.8} & 67.8\\
ORKG-TDM\textsubscript{Bert} & {68.1} & {67.5} & {65.5} & {79.6} & {63.3} & {70.5} \\
ORKG-TDM\textsubscript{SciBert} & {65.7} & {77.2} & {68.3} & {65.7} & {76.8} & {70.8} \\
ORKG-TDM\textsubscript{XLNet} & {71.7} & {73.9} & \textbf{70.6} & {77.1} & {70.9} & \textbf{73.9} \\
\hline 
\multicolumn{7}{c}{(b) Task + Dataset + Metric Extraction (without "Unknown" annotation)} \\
\hline 
TDM-IE\textsubscript{Bert} & 54.1 & {65.9} & 56.6 & 60.2 & {73.1} & 66.0 \\
ORKG-TDM\textsubscript{Bert} & {59.0} & {55.4} & {54.7} & {79.5} & {57.6} & {66.8} \\
ORKG-TDM\textsubscript{SciBert} & {57.6} & {68.7} & {60.1} & {65.3} & {73.1} & {69.0} \\
ORKG-TDM\textsubscript{XLNet} & {63.5} & {64.1} & \textbf{61.4} & {76.4} & {66.4} & \textbf{71.1} \\
\hline
\end{tabular}
\caption{\textsc{Leaderboard} triple extraction task, comparison of our ORKG-TDM models versus the original science result extractor model (first row in parts (a) and (b)) on the original corpus (see last two columns in \autoref{table:dataset_stats} for the original corpus statistics).}
\label{tab:exp}
\end{table}

\subsection{Experimental Results}
The results from our comprehensive evaluations with respect to four main research questions (noted as \textbf{RQ1}, \textbf{RQ2}, \textbf{RQ3}, and \textbf{RQ4}) are shown in Tables~\ref{tab:exp}, 3, and 4, respectively.

\subsubsection{RQ1: How well do our transformer models perform for \textsc{Leaderboard} extraction compared to the original science result extractor when trained in the identical original experimental setting?} In the last two columns in \autoref{table:dataset_stats}, we showed statistics of the comparatively smaller original corpus that defined and evaluated the \textsc{Leaderboard} extraction task~\cite{hou2019identification}. As a quick recap, the original corpus had 78 distinct TDM-triples including ``unknown,'' and a distribution of 170 papers in the train dataset and 167 papers in the test dataset as fixed partitions; there were 46 and 45 ``unknown'' papers in the train and test sets, respectively. We evaluate all the three transformer models on this original corpus to compare our model performances. These results are shown in \autoref{tab:exp}. As we can see in the table, all three of our models outperform the original with XLNet reporting the best score. We obtain a 70.6 macro F1 versus 65.3 in the baseline. The Bert model is only a few fractional points better at 65.5. Further, we obtain a 73.9 micro F1 versus 67.8 in the baseline, with the Bert model at 70.5. With these results our models outperform the original system in the original settings reported for this task.

Next, we examine the results of our models on our larger corpus for \textsc{Leaderboard} extraction.

\begin{table}[!tb]
\centering \scriptsize
\begin{tabular}{lcccccc}
\hline
\textbf{}        & \textbf{Macro P} & \textbf{Macro R} & \textbf{Macro F1} & \textbf{Micro P} & \textbf{Micro R} & \textbf{Micro F1} \\ \hline
                 & \multicolumn{6}{c}{Average Evaluation Accross 2-fold}                                                                  \\ \hline
ORKG-TDM\textsubscript{Bert} &   \textbf{92.8}        &  93.9        &    {92.4}             &      \textbf{95.5}      &     89.1      &     {92.1}       \\ \hline
ORKG-TDM\textsubscript{SciBert} &  90.9          &   93.4         &     91.1            &     94.1       &    88.5        &       91.2       \\ \hline
ORKG-TDM\textsubscript{XLNet} &  \textbf{92.8}            &     \textbf{94.8}       &     \textbf{92.8}          &     94.9       &    \textbf{91.2}        &     \textbf{93.0}     \\ \hline
& \multicolumn{6}{c}{Average Evaluation Accross 2-fold (without "Unknown" annotation)} \\
\hline 
ORKG-TDM\textsubscript{Bert} &     \textbf{91.7}        &      92.1       &      {90.8 }        &    \textbf{95.7}       &      88.3      &       {91.8}       \\ \hline
ORKG-TDM\textsubscript{SciBert} &   89.7           &   91.4         &     89.4           &      94.4       &    87.6        &       90.9      \\ \hline
ORKG-TDM\textsubscript{XLNet} &   91.6          &    \textbf{93.1}      &      \textbf{91.2 }       &    95.0        &    \textbf{90.5}         &     \textbf{92.7}        \\ \hline
\end{tabular}
\caption{Top results for Bert (10neg; 1,302unk), SciBert (10neg; 1,302unk), and XLNet (10neg; 1,302unk)}
\end{table}

\begin{table}[!tb]
\centering \scriptsize
\begin{tabular}{lcccccccccccc}
\hline\multirow{2}{*} {\textbf{Entity} } & \multicolumn{5}{c} { \textbf{Macro} } & & \multicolumn{5}{c} { \textbf{Micro} } \\
\cline { 2 - 6 } \cline { 8 - 12 } & $\mathrm{P}$ && $\mathrm{R}$ && $\mathrm{F}_{1}$ & & $\mathrm{P}$ && $\mathrm{R}$ && $\mathrm{F}_{1}$ \\
\hline TDM & 91.6 && 93.1 && 91.2 & & 95.0 && 90.5 && 92.7 \\
\hline
Task & \textbf{93.7} && \textbf{94.8} && \textbf{93.6} & & \textbf{97.4} && \textbf{93.6} && \textbf{95.5} \\
Dataset & \textbf{92.9} && 93.6 && {92.4} & & {96.6} && 91.5 && 94.0 \\
Metric & 92.5 && {94.2} && {92.5} & & 96.0 && 92.5 && \textbf{94.2} \\
\hline
\end{tabular}
\caption{Performance of our best model, i.e. ORKG-TDM\textsubscript{XLNet}, for \textit{Task}, \textit{Dataset}, and \textit{Metric} concept extraction of the \textsc{Leaderboard}}
\label{tab:t_d_m}
\end{table}

\begin{table}[!tb]
\centering
\resizebox{\textwidth}{!}{
\begin{tabular}{lcccccc}
\hline \textbf{Document Representation} & \textbf{Macro P} & \textbf{Macro R} & \textbf{Macro F1} & \textbf{Micro P} & \textbf{Micro R} & \textbf{Micro F1} \\
\hline Title + Abstract & 88.6 &	92.9 &	89.4 &	92.6&	90 & 91.3
 \\
Title + Abstract + ExpSetup & 89.2&	91.5&	89.2&	94.2&	89&	91.5
 \\
Title + Abstract + TableInfo & 90.5	&\textbf{94.4}&	91.2&	93.5&	93.2&	93.3 \\
Title + Abstract + ExpSetup + TableInfo & \textbf{92.3}&	93.5&	\textbf{91.7}&	\textbf{95.1}&	92&	\textbf{93.5}
 \\
\hline
\end{tabular}
}
\caption{Ablation results of our best model, i.e. ORKG-TDM\textsubscript{XLNet}, for \textsc{Leaderboard} extraction as $(task, dataset, metric)$ triples}
\label{tab:expAbl}
\end{table}

\subsubsection{RQ2: How do the transformer models perform on a large corpus for \textsc{Leaderboard} construction?} We examine this question in light of the results reported in Table 3. We find that again, consistent with the observations made in Table 2, XLNet outperforms Bert and SciBert. Note that we have leveraged XLNet with the limited context length of 512 tokens (by truncating parts of the context described in Section 6.1) and thus the potential of leveraging the full context in XLNet models remains untapped. Nevertheless, XLNet still distinguishes itself from the Bert models with its PLM language modeling objective versus Bert's MLM. This in practice has also shown to perform better~\cite{yang2019xlnet}. Thus XLNet with limited context encoding can still outperform the Bert models. Among all the models, we find SciBert shows a slightly lower performance compared to Bert. This is a slight variation on performance observations obtained in the original smaller dataset (results in Table 2) where Bert was slightly lower than SciBert. These results indicate that in smaller datasets, the SciBert model should be preferred since its underlying science-specific pretrained corpus would compensate for signal absences in the task training corpus. However, with larger training datasets for finetuning the Bert models, the underlying pretraining corpus domains show not to be critical to the overall model performance.

Further, we observe the macro and micro evaluations for all three systems (ORKG-TDM\textsubscript{Bert}, ORKG-TDM\textsubscript{SciBert}, ORKG-TDM\textsubscript{XLNet}) show evenly balanced, similar scores. This tells us that our models handles the majority and minority TDM classes evenly. Thus given our large training corpus, the transformers remain unaffected by the underlying dataset class distributions.

\subsubsection{RQ3: Which of the three \textsc{Leaderboard} concepts are easy or challenging to extract?}

As a fine-grained examination of our best model, i.e. ORKG-TDM\textsubscript{XLNet}, we examined its performance for extracting each of three concepts $(Task, Dataset, Metric)$ separately. These results are shown in \autoref{tab:t_d_m}. From the results, we observe that \textsc{Task} is the easiest concept to extract, followed by \textsc{Metric}, and then \textsc{Dataset}. We ascribe the low performance for extracting the \textsc{Dataset} concept due to the variability in its naming seen across papers even when referring to the same real-world entity. For example, the real-world dataset entity \textit{CIFAR-10} is labeled as \textit{CIFAR-10, 4000 Labels} in some papers and \textit{CIFAR-10, 250 Labels} in others. This phenomenon is less prevalent for \textsc{Task} and the \textsc{Metric} concepts. For example, the \textsc{Task} `Question Answering' is rarely referenced differently across papers addressing the task. Similarly, for \textsc{Metric}, \textsc{Accuracy}, as an example, has very few variables namings.

\subsubsection{RQ4: Which aspect of the DocTAET context feature representation had the highest impact for \textsc{Leaderboard} extraction?}

Further, in \autoref{tab:expAbl}, we breakdown performance of our best model, i.e. ORKG-TDM\textsubscript{XLNet}, examining the impact of the features as the shortened context from the articles for TDM inference. We observe, that adding additional contextual information in addition to title and abstract increases the performance significantly, while the actual type of additional information (i.e. experimental setup, table information or both) impacts the performance to a lower extend.

\subsection{Integrating ORKG-TDM in Scholarly Digital Libraries}

\begin{figure}[!tb]
	\centering
	\includegraphics[width=0.6\linewidth]{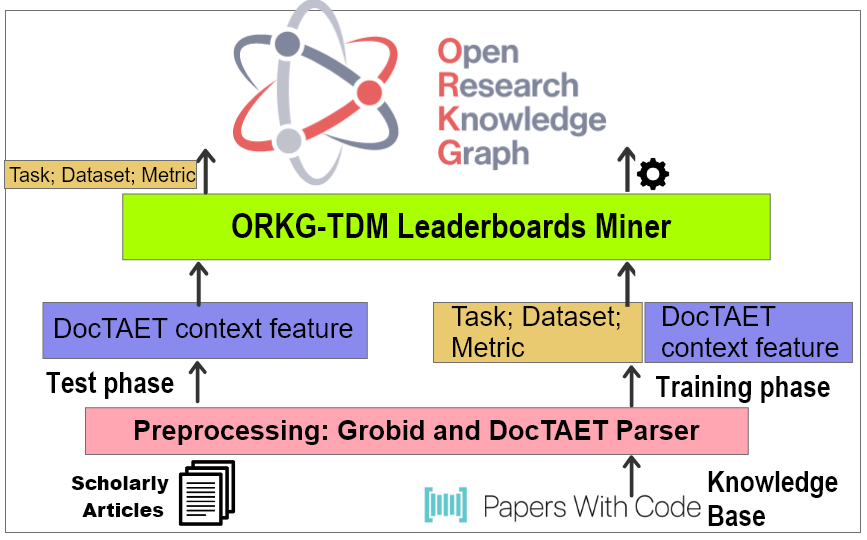}
	\caption{Overview of the \textsc{Leaderboard} Extraction Machine Learning Process Flow for ORKG-TDM in the Open Research Knowledge Graph Digital Library}
	\label{fig:sys}
\end{figure}

Ultimately, with \textsc{Leaderboard} construction we aim to give researchers oversight over the state-of-the-art wrt. certain research questions (i.e. tasks).
\autoref{fig:sys} shows how our ORKG-TDM \textsc{Leaderboard} mining is integrated into the Open Research Knowledge Graph scholarly knowledge platform.
The fact, that the automated \textsc{Leaderboard} mining results in incomplete and partially imprecise results can be alleviated by the crowdsourcing and curation features implemented in the ORKG.
Also, ORKG provides features for dynamic \textsc{Leaderboard} visualization, publication, versioning etc.

Although, the experiments of our study targeted empirical AI research, we are confident, that the approach is transferable to similar scholarly knowledge extraction tasks in other domains. For example in chemistry or material sciences, experimentally observed properties of substances or materials under certain conditions could be obtained from various papers.

\section{Conclusion and Future Work}
In this paper, we investigated the \textsc{Leaderboard} extraction task w.r.t. three different transformer-based models. Our overarching aim with this work is to build a system for comparable scientific concept extractors from scholarly articles. Therefore as a next step, we will extend the current triples (task, dataset, metric) model with additional concepts that are suitable candidates for a \textsc{Leaderboard} such as \textit{score} or \textit{code urls}, etc. In this respect, we will adopt a hybrid system wherein some elements will be extracted by the machine learning system as discussed in this work while other elements will be extracted by a system of rules and regular expressions.
Also, we plan to combine the automated techniques presented herein with a crowdsourcing approach for further validating the extracted results and providing additional training data. Our work in this regard is embedded in a larger research and service development agenda, where we build a comprehensive knowledge graph for representing and tracking scholarly advancements~\cite{jaradeh2019open}.
We also envision the task-dataset-metric extraction approach to be transferable to other domains (such as materials science, engineering simulations etc.).
Our ultimate target is to create a comprehensive structured knowledge graph tracking scientific progress in various scientific domains, which can be leveraged for novel machine-assistance measures in scholarly communication, such as question answering, faceted exploration and contribution correlation tracing.

\subsubsection*{Acknowledgements}
This work was co-funded by the Federal Ministry of Education and Research (BMBF) of Germany for the project LeibnizKILabor (grant no. 01DD20003) and by the European Research Council for the project ScienceGRAPH (Grant agreement ID: 819536).

%

%
%
%
%

\bibliographystyle{splncs04}
\bibliography{main}

\appendix

\section{Examples of Leaderboards in Our Corpus}

The top-3 most common \textsc{Leaderboard}s in our training set included: 1) \textit{Image Classification, ImageNet, Top 1 Accuracy}; 2) \textit{Object Detection, COCO test-dev, box AP}; 3) \textit{Image Classification, CIFAR-10, Percentage correct} occurring 93, 57, and 51 times, respectively.

The top-3 least common randomly selected \textsc{Leaderboard}s in our training set included: 1) \textit{Word Sense Disambiguation, WiC-TSV, Task 1 Accuracy: all}; 2) \textit{Entity Linking, WiC-TSV, Task 1 Accuracy: all}; 3) \textit{Causal Inference, IDHP, Average Treatment Effect Error} occurring once.

\end{document}